\documentclass{article}
\usepackage{arxiv}
\usepackage[T1]{fontenc}
\usepackage{graphicx,amsmath,amssymb,booktabs,longtable,array,caption,float,placeins,url,xurl}
\usepackage{microtype,enumitem,tabularx}
\usepackage[natbibapa]{apacite}
\usepackage[hypertexnames=false]{hyperref}
\AtBeginDocument{}
\hypersetup{colorlinks=true,linkcolor=black,citecolor=black,urlcolor=blue,
  pdftitle={Seeing Is No Longer Believing: Frontier Image Generation Models, Synthetic Visual Evidence, and Real-World Risk},
  pdfauthor={Shuai Wu, Xue Li, Zhijun Wang, Bolun Liu, Weilin Cai, Zihao Su, Ran Wang},
  pdfsubject={Research preprint}}
\newcommand{\modelname}[1]{\textsc{#1}}
\newcolumntype{Y}{>{\raggedright\arraybackslash}X}
\setlist[itemize]{leftmargin=*}
\setlist[enumerate]{leftmargin=*}

\title{Seeing Is No Longer Believing:\\Frontier Image Generation Models,\\Synthetic Visual Evidence, and Real-World Risk}
\author{
  Shuai Wu\thanks{Corresponding author: \href{mailto:shuai.wu@colorado.edu}{shuai.wu@colorado.edu}.\newline ORCID: Shuai Wu: \href{https://orcid.org/0009-0007-7657-6208}{0009-0007-7657-6208}; Xue Li: \href{https://orcid.org/0009-0002-7480-723X}{0009-0002-7480-723X}; Zhijun Wang: \href{https://orcid.org/0009-0000-0742-6985}{0009-0000-0742-6985}.}\\
  \normalfont\footnotesize University of Colorado Boulder
  \And
  Xue Li\\
  \normalfont\footnotesize Beijing Union University
  \And
  Zhijun Wang\\
  \normalfont\footnotesize University of North Dakota
  \And
  Bolun Liu\\
  \normalfont\footnotesize City University of Macau
  \AND
  Weilin Cai\\
  \normalfont\footnotesize The University of Sydney
  \And
  Zihao Su\\
  \normalfont\footnotesize Beijing Information Science\\
  \normalfont\footnotesize and Technology University
  \And
  Ran Wang\\
  \normalfont\footnotesize Taiyuan University of Technology
}
\date{October 2026}
\renewcommand{\shorttitle}{Seeing Is No Longer Believing}
\begin{document}
\shortcites{ho2020ddpm,rombach2022latent,saharia2022imagen,peebles2023dit,tordjman2026xrays,cao2026marknull}
\maketitle

\begin{abstract}
Image generation systems can produce plausible photographs, readable documents, and consistent depictions of people and places. Presented as records of real events, these outputs can influence decisions in news, finance, identity verification, medicine, and law. This narrative review examines public model documentation, incident reports, empirical studies, governance materials, and illustrative English and Chinese practitioner commentary available through 1 October 2026. We separate vendor capability claims, documented incidents, experimental findings, and prospective harm pathways. Realism, text rendering, reference consistency, conversational editing, search grounding, and low generation cost shape the conditions under which synthetic images acquire evidentiary weight. Documented episodes show how fabricated visuals enter public and institutional workflows, although public reporting cannot measure misuse prevalence for current releases. Comparing provider safeguards, provenance credentials, watermarks, platform labels, and statutory duties shows that origin signals do not verify a depicted event or transaction. Our framework matches artifact types and decision stakes to specific controls. Consequential decisions require authenticated source records, independent channel corroboration, and proportionate review before action.
\end{abstract}

\section{Introduction}
\label{sec:intro}

Photographs often carry evidentiary weight because viewers associate them with optical capture. Staging, darkroom or digital editing, and misleading captions have long complicated that assumption. Current image generators offer another route to a plausible visual record by letting users specify a scene or document in natural language and refine it across conversational turns. Once these outputs enter social feeds or institutional workflows, recipients may treat visual coherence as proof of the depicted event or transaction.

This narrative review examines commercial and open-weight image generators documented through 1 October 2026 and analyzes how their capabilities affect visual verification. Reviewed releases include \modelname{GPT Image 2.5}, Google's \modelname{Nano Banana} family, \modelname{Grok Imagine Image 2.0}, \modelname{Qwen-Image 3.0 Pro}, \modelname{Qwen-Image 2.1}, and \modelname{Seedream 5.0}. Primary vendor sources and additional commercial or open-weight models appear in Table~\ref{tab:models}. Across these systems, vendor documentation describes photorealistic rendering, multilingual typography, reference-image conditioning, localized editing, search grounding, and provenance metadata. We examine how these documented functions bear on institutional verification workflows.

Public reporting records several distinct forms of synthetic visual deception. In April 2026, Chinese press reports examined fabricated corporate notices and conversational edits to identity-document images \citep{thepaper2026gptimage,jiemian2026gptimage}. In 2026, Google withdrew an image-generation tool from Google Earth after misleading reconstructions spread from its mapping interface \citep{google2026earth,pcmag2026nbge}, and the \citet{opc2026grok} found consent violations in non-consensual sexualized deepfakes generated with Grok. Earlier fact-checks debunked a false Pentagon explosion and synthetic scenes of aircraft landing at a burning Beirut airport \citep{ap2023pentagon,reuters2024beirut}, alongside fabricated photographs of Donald Trump's arrest and Pope Francis in a puffer jacket \citep{ap2023trumparrest,cbs2023pope}. Controlled radiograph reader experiments and commercial identity-fraud statistics add evidence within narrower study and service boundaries \citep{tordjman2026xrays,entrust2025fraud}.

Our analysis links documented generation features to plausible misuse pathways and institutional verification duties. Historical cases illustrate artifact forms and circulation channels, and we attribute an episode to a specific generator only when the cited source confirms it. Combining case evidence with a decision-centered framework clarifies how synthetic images acquire an evidentiary role.

\begin{figure}[!htbp]
    \centering
    \includegraphics[width=0.98\textwidth]{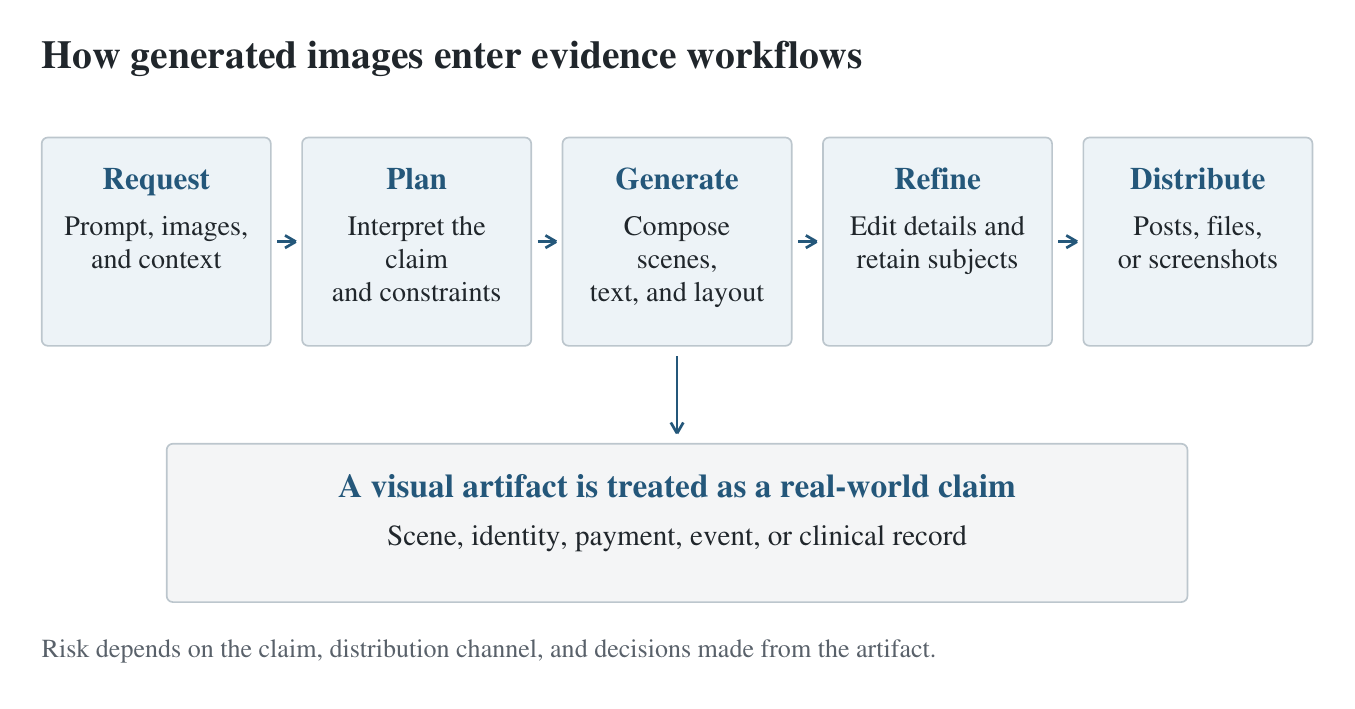}
    \caption{Pathway from a generation prompt to an image treated as evidence. Stages represent functional steps across generation and distribution workflows rather than a uniform architecture shared by every model.}
    \label{fig:pipeline}
\end{figure}

\subsection{Contributions}
This review makes three contributions:
\begin{enumerate}
  \item It separates vendor capability claims through 1 October 2026 from documented incidents and empirical studies across hosted and open-weight image generators.
  \item It explains how readable typography, identity persistence, search grounding, iterative editing, and low marginal cost lend synthetic images apparent evidentiary authority.
  \item It distinguishes file-origin provenance such as C2PA manifests and watermarks from independent corroboration across seven high-stakes decision domains.
\end{enumerate}

\section{Related Work}
\label{sec:related}

\subsection{Generative image models}
Diffusion architectures underpin most contemporary image generators. \citet{ho2020ddpm} introduced denoising diffusion probabilistic models that invert a progressive noise process, and \citet{rombach2022latent} moved denoising into a learned latent space to lower compute cost relative to pixel-space sampling. \citet{saharia2022imagen} coupled pretrained language encoders with text-conditioned synthesis, and \citet{peebles2023dit} replaced convolutional backbones with scalable diffusion transformers. Current commercial and open-weight releases pair visual synthesis with reference-image conditioning, multi-turn editing, search retrieval, and text layout, though architectural details and safety filters vary across products.

\subsection{Synthetic media, deepfakes, and visual misinformation}
Empirical research on synthetic media evaluates both visual discrimination and perceived credibility. \citet{nightingale2022faces} found that experimental participants performed near chance when distinguishing synthetic faces from real portraits and rated synthetic faces as more trustworthy. Newsroom fact-checks complement controlled studies with public cases: a fabricated 2023 Pentagon explosion circulated as breaking crisis coverage \citep{ap2023pentagon}, and synthetic group portraits linked Zohran Mamdani and other public figures to Jeffrey Epstein \citep{reuters2026mamdani,reuters2026epstein}. These studies and fact-checks show how visual plausibility, publication framing, and external corroboration shape the treatment of images as proof.

\subsection{Provenance and authenticity infrastructure}
Provenance credentials and post-hoc detectors answer different forensic questions. C2PA manifests bind cryptographically signed assertions about file creation and subsequent edits to media assets \citep{c2pa2026assertions}. Valid credentials can document how a file was produced, yet they do not verify whether a depicted event occurred. Platform re-encoding, file conversion, or screen capture can strip embedded manifests. Invisible watermarks embed an alternative origin signal into pixel values for recovery by compatible detectors. In a case study of Adobe's deployment, the \citet{pai2024adobe} examines how provenance metadata and user-facing disclosures fit within media-stewardship workflows alongside source checks and editorial review.

\section{Current Image Generation Models}
\label{sec:models}

Table~\ref{tab:models} summarizes public capability claims relevant to synthetic visual evidence for model releases available on 1 October 2026. Each row pairs vendor-documented specifications with their analytical implications for evidentiary workflows.

{\small
\setlength{\tabcolsep}{4pt}
\begin{longtable}{>{\raggedright\arraybackslash}p{0.18\textwidth}>{\raggedright\arraybackslash}p{0.12\textwidth}>{\raggedright\arraybackslash}p{0.35\textwidth}>{\raggedright\arraybackslash}p{0.275\textwidth}}
\caption{Image generation models and documented capabilities as of 1 October 2026.}\label{tab:models}\\
\toprule
\textbf{Model} & \textbf{Developer} & \textbf{Publicly described capability} & \textbf{Relevance to visual evidence} \\
\midrule
\endfirsthead
\multicolumn{4}{l}{\textbf{Table \thetable\ continued.}}\\
\toprule
\textbf{Model} & \textbf{Developer} & \textbf{Publicly described capability} & \textbf{Relevance to visual evidence} \\
\midrule
\endhead
\midrule\multicolumn{4}{r}{Continued on next page}\\\endfoot
\bottomrule\endlastfoot
GPT Image 2.5 / ChatGPT Images 2.5 & OpenAI & Release notes describe tighter instruction following, multilingual text rendering, and iterative editing \citep{openai2026images25}, paired with C2PA credentials and SynthID provenance \citep{openai2026provenance}. & Readable text and localized edits can produce plausible notices or interface screenshots as candidate forgeries. \\
Nano Banana Pro & Google & Gemini 3 Pro Image supports up to 4K output, fourteen reference images, five consistent characters, Google Search grounding, and provenance signals \citep{google2025nanobananapro,googlecloud2026ga}. & Cross-image identity consistency and grounded details can make multiple fabricated images appear mutually corroborating. \\
Nano Banana 2 & Google & Gemini 3.1 Flash Image pairs low-latency generation with subject and object consistency, image-search grounding, and SynthID watermarking \citep{google2026nanobanana2}. & Rapid turnaround and multi-turn edits lower the effort required to assemble coherent visual narratives. \\
Nano Banana 2 Lite & Google & Gemini 3.1 Flash-Lite Image targets high-throughput 1K generation at reduced latency and per-image cost \citep{google2026nanobanana2lite}. & Low API pricing reduces the marginal cost of iterative drafting and automated image pipelines. \\
Grok Imagine Image 2.0 & xAI & Supports generation, localized editing, planned text layouts, aspect-ratio resizing, reference inputs, and batch API processing \citep{xai2026imagine20,xai2026imagegeneration}. & Layout planning and batch endpoints facilitate repeated production of formatted visuals. \\
Qwen-Image 3.0 Pro & Alibaba & Hosted service combining generation and instruction editing with one to three reference images and configurable resolutions \citep{qwen2026image30,qwen2026api30}. & Dense multilingual typography supports document-style outputs under hosted reference limits. \\
Qwen-Image 2.1 & Alibaba & Open-weight release supporting up to ten reference images, local mask editing, and RGBA transparency \citep{qwen2026image21}. & Local inference operates outside hosted account moderation under the release license and applicable law. \\
Seedream 5.0 family & ByteDance & Lite, Pro, and Flash tiers combine generation, region editing, contextual retrieval, and high-volume output \citep{bytedance2026seedream,picsart2026seedream}. & Region-level photo edits and grounded details can reinforce altered records across distinct price tiers. \\
Additional commercial and open-weight releases & Multiple & Commercial releases include Midjourney V8.2, FLUX 3, and Recraft V4.1 \citep{midjourney2026v82,bfl2026flux3,recraft2026v41}, alongside Kling 3.0 Omni and open-weight HunyuanImage-3.0 \citep{aliyun2026kling,tencent2025hunyuan30}. Ideogram 4.0 distributes weights under a Non-Commercial License and inference code under Apache 2.0 \citep{ideogram2026v40,ideogram2026license}. & Governance analysis must separate hosted interface controls, weight availability, and distinct code versus model licenses. \\
\end{longtable}
\par\footnotesize\textit{Note.} Capability descriptions reflect public vendor documentation available on 1 October 2026. Evidentiary implications represent analytical inferences rather than measured fraud rates across an illustrative subset of releases.
}

\begin{figure}[!htbp]
    \centering
    \includegraphics[width=0.96\textwidth]{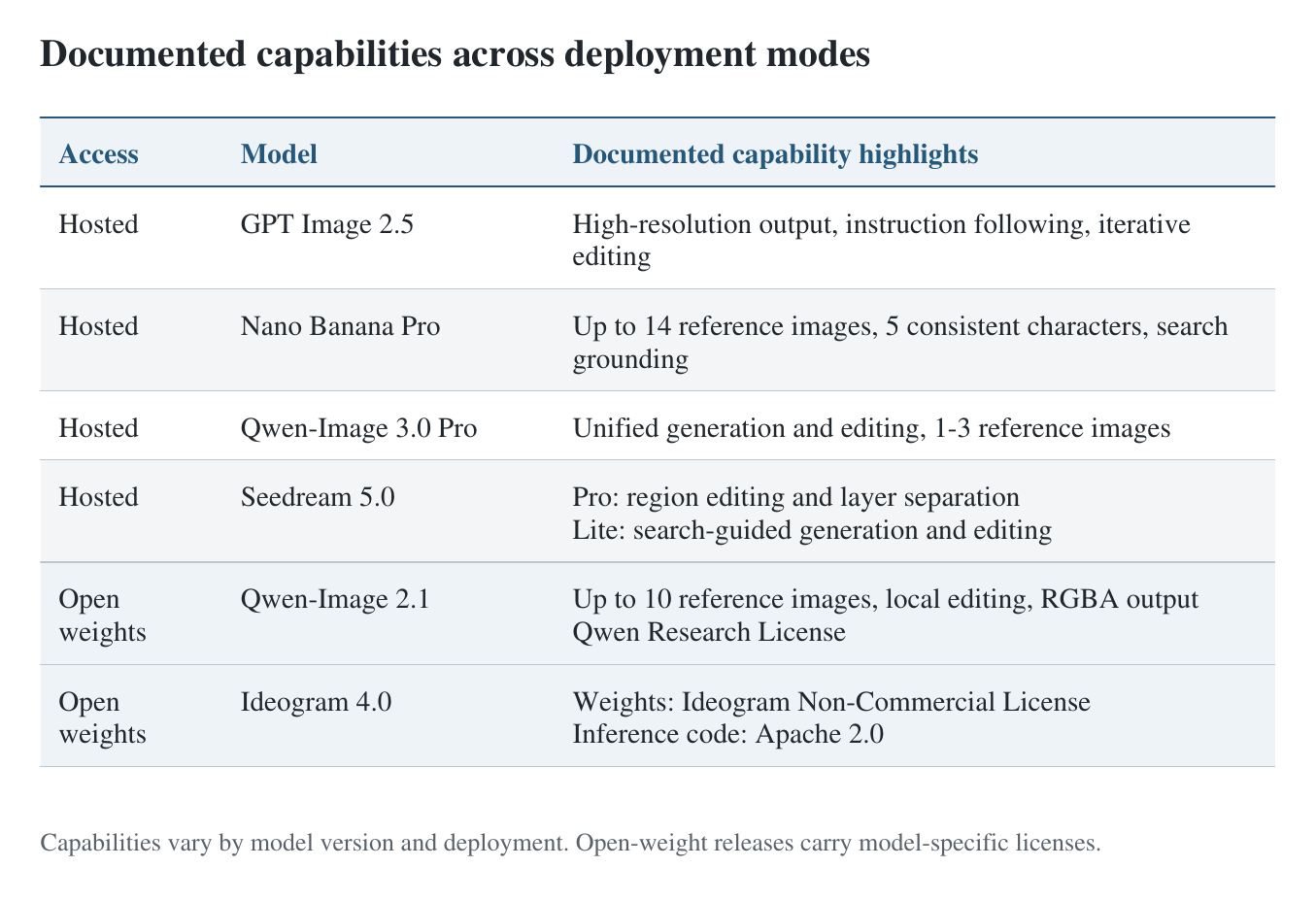}
    \caption{Documented capabilities across hosted and open-weight releases. Table~\ref{tab:models} lists primary sources and version scope for each model.}
    \label{fig:capability_heatmap}
\end{figure}

\subsection{Deployment choices and capability claims}
Reviewed releases span high-control studio tiers, low-cost drafting endpoints, batch APIs, and downloadable weights. Google documents web or image-search grounding for Nano Banana Pro and Nano Banana 2 \citep{googlecloud2026ga,google2026nanobanana2}. Alibaba separates hosted Qwen-Image 3.0 Pro from open-weight Qwen-Image 2.1 in reference-image capacity and RGBA output \citep{qwen2026api30,qwen2026image21}. Deployment mode determines where safeguards can operate. Open weights allow local configuration outside hosted moderation, leaving governance to model licenses, hardware constraints, and legal rules on downstream distribution \citep{ideogram2026license}. Vendor specifications support feature-level comparison, and head-to-head reliability or safety claims require shared evaluation benchmarks.

\section{Technical Background: From Image Synthesis to Synthetic Evidence}
\label{sec:technical}

\subsection{Prompt-conditioned visual construction}
Modern image-generation interfaces accept natural-language prompts specifying subjects, settings, camera treatment, typography, and page structure. Both hosted services and open-weight releases also accept reference images or edit uploaded files directly \citep{openai2026images25,qwen2026image21}. Such controls let users assemble outputs that resemble either optical photographs or formatted records. Evidentiary plausibility depends on internal coherence. A fabricated notice gains credibility when its wording matches institutional layout conventions, and a synthetic event photo appears more convincing when scene details align with its caption. Task-specific testing is needed to determine how reliably a given model maintains that coherence.

\subsection{Text rendering as a risk amplifier}
Accurate text rendering expands synthesis from pictorial scenes to textual records. Vendor announcements for ChatGPT Images 2.0 and Qwen-Image 3.0 describe multilingual lettering, posters, infographics, and document-style page layouts \citep{openai2026images2,qwen2026image30}, and Nano Banana Pro extends typography to 4K studio compositions \citep{google2025nanobananapro}. Those capabilities support graphic design and localization, yet they also supply the visual grammar of counterfeit notices, invoices, receipts, and interface screenshots. Risk assessment asks whether generated text is legible and consistent enough to pass initial review in a target workflow, because rendering accuracy varies across scripts, font sizes, and document layouts.

\subsection{Reference consistency and identity persistence}
Reference-image conditioning preserves facial identity, objects, or graphic motifs across multiple outputs. Documented input budgets vary by model. Nano Banana Pro accepts up to fourteen reference images and maintains five consistent characters \citep{google2025nanobananapro}. Hosted Qwen-Image 3.0 Pro accepts one to three reference images \citep{qwen2026api30}, and open-weight Qwen-Image 2.1 accepts up to ten editing inputs with RGBA output \citep{qwen2026image21}. Storyboarding and brand design rely on this persistence. In adversarial settings, shared reference inputs allow an operator to place the same real person or document template across a series of fabricated scenes. Visual agreement across images derived from common references therefore provides weak independent corroboration.

\subsection{Reasoning and grounding}
GPT Image 2.5 and Seedream 5.0 Lite incorporate pre-generation reasoning or live web retrieval \citep{openai2026images25,bytedance2026seedream}, and Nano Banana 2 adds image-search grounding \citep{google2026nanobanana2}. Grounding helps a model incorporate accurate landmarks, weather conditions, or institutional branding into a composition. Embedded inside a fabricated event, accurate contextual details can make a false claim harder to spot on casual inspection. Search retrieval may render a real street sign or corporate emblem inside an incident that never happened. Verification workflows must therefore separate the accuracy of background details from the truth of the asserted event.

\subsection{Editing and iterative repair}
Multi-turn conversational editing lets users refine specific regions and preserve surrounding visual details \citep{openai2026images25}. An adversary can use sequential edits to fix conspicuous rendering flaws or tailor an existing photograph or document to a target intake process. Operational effectiveness depends on both visual stability across rounds and how refusal filters respond across a sequence of revision prompts. Security evaluations should therefore test multi-turn editing trajectories rather than single prompts in isolation.

\subsection{Price, latency, and volume}
Latency and per-image pricing govern how many candidate outputs an operator can generate and screen. Nano Banana 2 Lite and Seedream 5.0 offer low-latency, lower-priced hosted endpoints for high-volume workflows \citep{google2026nanobanana2lite,bytedance2026seedream}. Local open-weight inference removes per-call API charges and shifts expenses to GPU hardware, electricity, and maintenance. Lower marginal cost makes trial-and-error refinement more affordable, even though published price schedules measure access conditions rather than observed abuse volume.

\section{Methodology}
\label{sec:method}
\subsection{Scope and source selection}
We conduct a narrative review of public sources to examine how synthetic images enter evidentiary workflows and how technical and institutional controls respond. Materials were collected through 1 October 2026. Product specifications and regulatory texts reflect the model versions and legal instruments active by that cutoff. Earlier empirical studies and landmark incidents are included where they clarify specific harm pathways. Illustrative English and Chinese community commentary spans November 2025 through 1 October 2026. The accompanying research notes organize that commentary by model family and theme as qualitative context rather than a complete query archive or probability sample.

Included sources address model capabilities, documented incidents, experimental evaluations, or governance mechanisms. Vendor pages and API guides supply product specifications. Peer-reviewed and preprint studies provide experimental measurements under controlled protocols. Statutes, regulatory findings, and technical standards define legal duties and provenance formats, and newsroom investigations or fact-checks document public circulation episodes. Practitioner threads and product reviews add user-reported observations, with attribution to provider claims, third-party estimates, or individual posts preserved throughout.

\subsection{Evidence synthesis}
Each source is analyzed across six questions covering the capability described, the artifact produced, the decision exposed, the harm mechanism observed or proposed, the applicable control, and remaining uncertainty. An incident report documents a concrete circulation episode, whereas a controlled experiment measures human or detector performance under its stated protocol. Vendor capability descriptions identify plausible threat pathways when an artifact enters a specified institutional workflow. Keeping these evidence classes distinct prevents capability announcements or isolated hoaxes from being misread as prevalence estimates.

\begin{figure}[!htbp]
    \centering
    \includegraphics[width=0.88\textwidth]{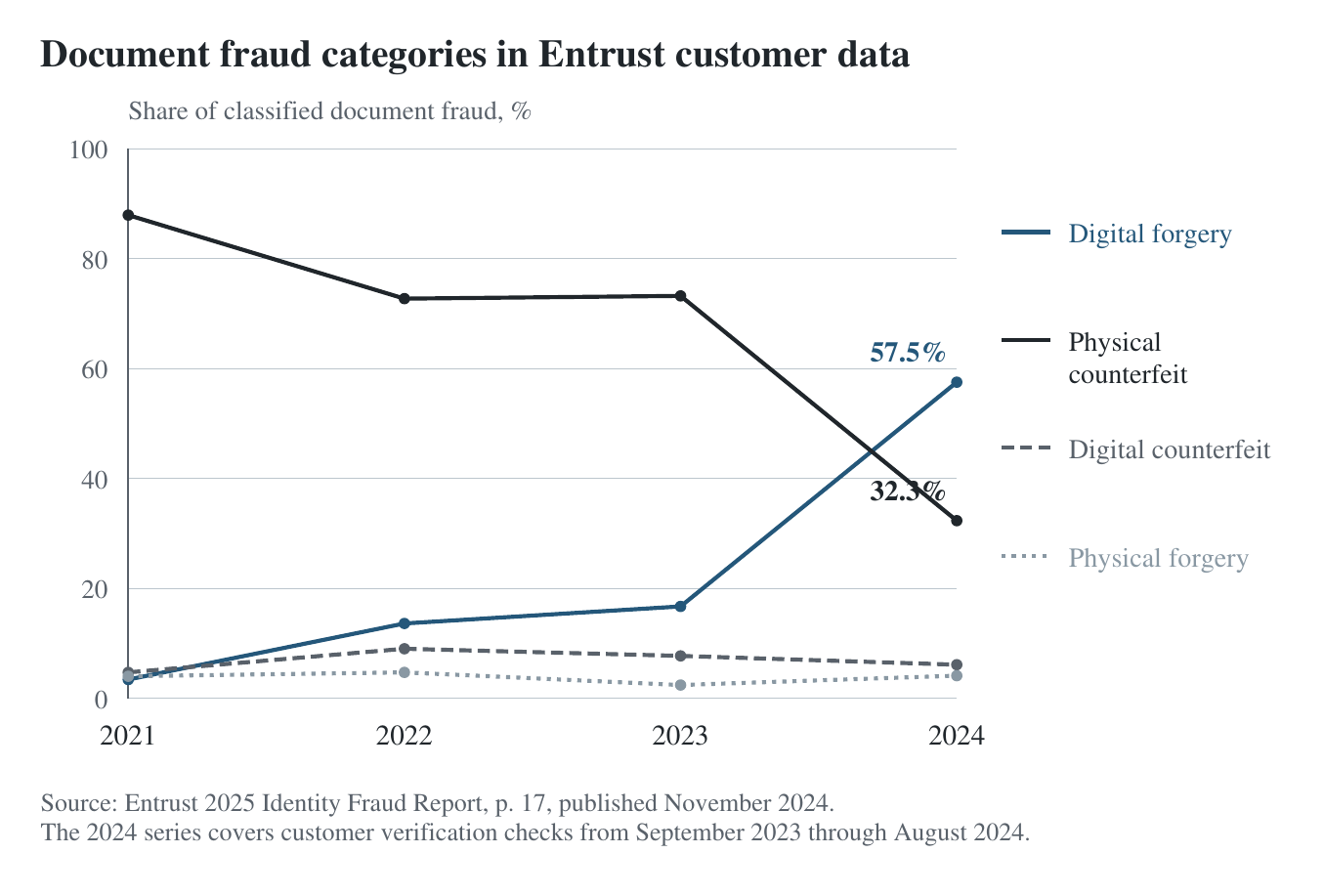}
    \caption{Category shares of classified document fraud in customer verification checks reported by \citet{entrust2025fraud} in November 2024. Annual labels follow the report series, where the 2024 window spans September 2023 through August 2024 within Entrust's customer population rather than global fraud prevalence.}
    \label{fig:document_trend}
\end{figure}

Figures summarize conceptual pathways, documented model features, and statutory timelines. Figure~\ref{fig:document_trend} plots fraud-category shares reported by \citet{entrust2025fraud} within its customer verification sample and observation window. None of the figures estimates population-wide abuse prevalence, cross-model safety rankings, or sector-level harm probabilities. Table~\ref{tab:evidence_types} defines the interpretive scope of each source class.

{\small
\begin{longtable}{>{\raggedright\arraybackslash}p{0.23\textwidth}>{\raggedright\arraybackslash}p{0.34\textwidth}>{\raggedright\arraybackslash}p{0.34\textwidth}}
\caption{Evidence types and their interpretive scope in this review.}\label{tab:evidence_types}\\
\toprule\textbf{Source type} & \textbf{Supported interpretation} & \textbf{Boundary}\\\midrule
\endfirsthead
\multicolumn{3}{l}{\textbf{Table \thetable\ continued.}}\\
\toprule\textbf{Source type} & \textbf{Supported interpretation} & \textbf{Boundary}\\\midrule\endhead
\bottomrule\endlastfoot
Provider documentation & Publicly stated feature, limit, or safeguard for a named release. & Real-world performance and policy enforcement vary by version, prompt, and deployment.\\
Incident reporting or fact-check & Verified circulation episode, debunked artifact, or institutional response. & Case reports do not measure overall incident frequency or identify the generator unless confirmed.\\
Original experimental research & Measured human or technical performance under a stated study protocol. & Generalization to other models, modalities, or operational workflows requires separate testing.\\
Regulatory or standards material & Statutory duty, enforcement finding, or technical specification from an authoritative body. & Jurisdictional reach, effective dates, and voluntary versus binding terms require separate analysis.\\
Community material & User-reported workflows, failure modes, or deployment practices. & Self-selection, unverified claims, and uneven platform retrieval prevent population inference.\\
\end{longtable}
}

\subsection{Conceptual risk framework}
Our framework evaluates a synthetic image jointly with the decision workflow that receives it. Photorealism, legible text, identity consistency, search grounding, iterative editing, and low unit cost raise either visual plausibility or production throughput. Downstream harm depends on the action requested, the authority attributed to the sender, propagation speed, and the verification steps preceding action. A fabricated payment screenshot causes loss when an accounting team accepts the image as settlement authorization instead of checking transaction records. A synthetic disaster photo causes disruption when audiences act before geolocation and local source confirmation.

Accordingly, the framework maps model capabilities to workflow failure points and matching controls rather than a single composite risk score. Public sources lack shared tasks, calibrated weights, representative exposure rates, and comparable outcome metrics needed for numerical aggregation. Control selection follows the failure mechanism and the stakes of an erroneous decision, and requires evaluation within the deploying institution.

\section{Public Evidence and Case Studies}
\label{sec:cases}

\begin{figure}[!htbp]
    \centering
    \includegraphics[width=0.98\textwidth]{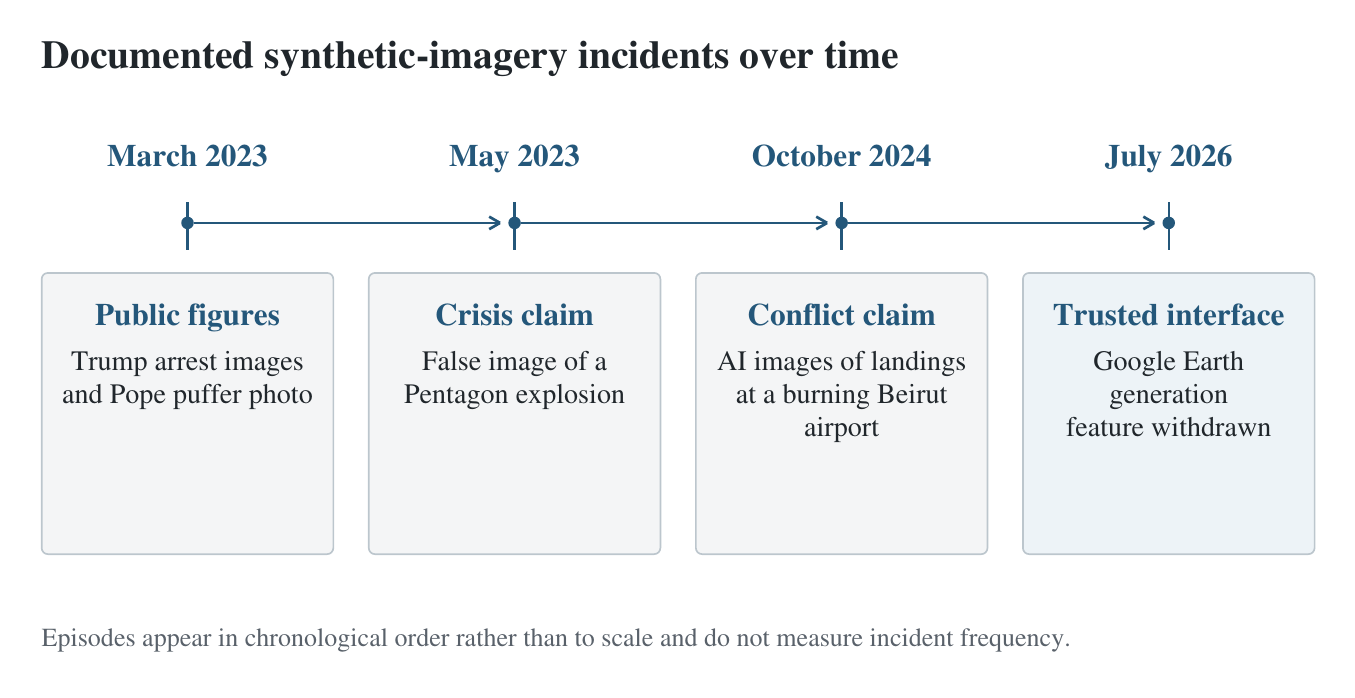}
    \caption{Chronological examples of publicly documented synthetic-imagery episodes from March 2023 through July 2026. Sources for each episode appear in Section~\ref{sec:cases}. Spacing is schematic and does not measure incident frequency or attribute older cases to current releases.}
    \label{fig:timeline}
\end{figure}

\subsection{False events and emergency news}
Fabricated crisis imagery can attract public or market reactions before authorities and newsrooms verify events on the ground. In May 2023, a synthetic image of an explosion near the Pentagon drew official denials after coinciding with a brief equity-market dip \citep{ap2023pentagon,reuters2023pentagon}. \citet{reuters2024beirut} later debunked AI-generated photographs purporting to show commercial aircraft landing at a burning Beirut airport. In July 2026, Google launched and withdrew an experimental Google Earth generation tool within one day \citep{google2026earth}. Press tests and coverage showed how misleading reconstructions could circulate from a familiar mapping interface \citep{pcmag2026nbge,zdnet2026ge}. \citet{journal2026garvaghy} also documented synthetic imagery circulating online during a confrontation in Northern Ireland. Across differing origins and consequences, each episode presented a synthetic visual as contemporaneous proof of a public event.

During the 2026 World Cup, the \citet{ebu2026worldcup} tracked viral posts blending synthetic visuals with false tournament claims. \citet{cq2026labelcase} reported an administrative detention for the publisher of a fabricated Chongqing bridge-collapse video even though the post carried an AI-generated label. That enforcement action separates origin disclosure from substantive falsity: a production label does not shield a fabricated emergency claim from verification or legal liability.

Recognizable local details, legible signs, and familiar platform layouts can heighten the apparent credibility of an emergency image. Urgent captions then encourage rapid sharing before verification. Case reports suggest this mechanism qualitatively rather than quantifying how model features alter audience behavior. Seeking confirmation from emergency authorities or an independently contacted local source remains a practical safeguard before acting on crisis imagery.

\subsection{Corporate notices and administrative trust}
In April 2026, Chinese press reports examined fabricated corporate announcements such as a false executive appointment poster and a forged company-dissolution statement \citep{thepaper2026gptimage,jiemian2026gptimage}. Both outlets also tested conversational edits that replaced portraits, names, and dates on identity-document images. Those reports show how multilingual text rendering and layout mimicry extend synthetic media into administrative records, although journalistic tests do not measure forgery prevalence or model reliability across document classes.

Institutional formatting can make a fabricated notice look routine on initial inspection. Receiving organizations should verify the issuing source, document identifier, and underlying record through an authenticated channel rather than treating logos, seals, or reference numbers inside the image as self-authenticating.

\subsection{Financial markets, invoices, and institutional workflows}
Visual misinformation affects finance through both public market reactions to crisis hoaxes \citep{ap2023pentagon} and payment fraud inside organizational workflows. Guidance from the \citet{icaew2026invoice} describes how invoice fraud pairs compromised vendor correspondence with altered payment instructions. Image generators could supply realistic invoices, payment screenshots, or supporting attachments within such schemes. Published invoice-fraud advisories describe the workflow mechanism and do not attribute documented cases to a named image model.

In its November 2024 \textit{2025 Identity Fraud Report}, \citet{entrust2025fraud} analyzed customer identity-verification checks from 1 September 2023 through 31 August 2024. Digital forgeries accounted for 57.46\% of detected document fraud in that dataset. Because that share reflects fraud caught within one commercial verification service without isolating generative models, it characterizes the report sample and points to growing operational pressure from digital document tampering.

Adjacent synthetic-media fraud shows how multiple communication channels can reinforce an apparently authoritative payment demand. \citet{gulfnews2026fideuram} reported a scheme targeting the Italian wealth manager Fideuram through cloned executive voices and impersonated messaging threads. That episode provides multi-channel context for visual-document controls without attributing the reported transfers to image generation.

\subsection{Medicine and clinical evidence}
\citet{tordjman2026xrays} examined synthetic radiographs in a controlled reader experiment with seventeen radiologists reviewing two non-overlapping sets totaling 264 images, half authentic and half synthetic. One set used ChatGPT-4o images across anatomical regions, and the other used RoentGen chest radiographs. Before being informed of the study's true purpose, 41\% of readers spontaneously identified the ChatGPT-4o outputs as AI-generated. Once told that synthetic images were present, readers reached a mean discrimination accuracy of 75\% on the ChatGPT-4o task. These results show a perceptual verification gap under experimental reading conditions, yet they do not document altered patient care or measure clinical deception rates.

Clinical exposure arises when an image reaches a clinician or payer through an unverified patient upload, external referral, or message. Hospitals and clinics can reduce that exposure by tying diagnostic images to authenticated acquisition devices, electronic health records, and Picture Archiving and Communication System logs. Source verification and clinical correlation against patient history should precede decisions based on externally supplied images.

\subsection{Public figures, celebrities, and the social proof problem}
Fact-checkers have documented synthetic photographs placing public figures in fabricated scenes. Prominent examples include synthetic arrest images of Donald Trump and photographs of Pope Francis in a puffer jacket \citep{ap2023trumparrest,cbs2023pope}, as well as invented group portraits linking Zohran Mamdani or other public figures to Jeffrey Epstein \citep{reuters2026mamdani,reuters2026epstein}. Placing a recognizable identity inside a staged encounter manufactures visual social proof around a reputational claim.

On 11 June 2026, the \citet{opc2026grok} concluded that X Corp. and xAI failed to obtain valid consent for collecting, using, and disclosing personal information to generate sexualized deepfakes. Its report cited news estimates of approximately 1.8 million sexualized images generated since 29 December 2025 and more than 6,000 per hour by 8 January 2026, and the Commissioner declined to determine an exact volume. Both companies committed to quarterly reporting and annual independent third-party audits. The Commissioner recorded those commitments as follow-up measures and reserved judgment on their effectiveness.

Recent political hoaxes illustrate complementary verification routes. \citet{washingtonpost2026trumpharp} reported the circulation of a fabricated photograph depicting Donald Trump kissing an aide, and \citet{techcrunch2026mcconnell} reported how Google's watermark detector helped debunk a hoax photograph of a hospitalized senator. Fact-checking addresses the truth of the depicted encounter, whereas a watermark check identifies output from a participating generator and leaves unwatermarked models or stripped copies to separate review.

Visual verification must account for both false acceptance and false rejection. Viewers may accept a fabricated image as authentic or dismiss genuine visual evidence out of concern over synthetic manipulation. Independent source records and corroborating evidence help resolve disputed claims in either direction.

\subsection{Documents, seals, contracts, e-mail, and screenshots}
Document-style generation extends evidentiary risk beyond photography to invoices, receipts, contracts, notices, badges, and interface screenshots. Controlled reporting by \citet{reuters2025phishingbots} showed that chatbots assisted with convincing phishing text. Document-capable image models allow comparable institutional phrasing to appear inside plausible visual layouts, creating an analytical risk pathway for document workflows.

Verification rigor should match the decision supported by an administrative artifact. Accounting staff can reconcile a transfer screenshot against ledger records. Legal counsel can check a contract image against the executed instrument, and recipients of a notice can confirm its validity with the issuing institution.

\section{Community and Practitioner Discussion Across Channels}
\label{sec:community}

Public commentary surfaces operational questions that practitioners and viewers raise around synthetic imagery. Drawing on linked English and Chinese sources, this section separates vendor specifications, journalist reviews, and individual user posts as illustrative signals rather than a representative sample or longitudinal survey.

\subsection{English-language public discussion}
In a Reddit discussion of YouTube's May 2026 automatic-labeling announcement, contributors raised concerns over false labels on conventional edits, missed synthetic videos, creator appeal procedures, and viewer filtering controls \citep{reddit2026youtubelabels}. Those comments identify concrete design criteria for evaluating label precision, appeal workflows, and viewer settings.

\subsection{Chinese-language reporting and commentary}
In April 2026, \citet{thepaper2026gptimage} and \citet{jiemian2026gptimage} combined hands-on editing tests of fabricated notices, identity documents, and chat screenshots with commentary on image credibility and China's synthetic-content labeling rules. Their reporting illustrates Chinese public debate over administrative trust and motivates out-of-band source checks for digital records.

\subsection{Review media}
Technology reviews examine distinct product trade-offs. On TechRadar, \citet{techradar2026realorai} published a real-or-AI reader quiz to illustrate perceptual ambiguity, although an informal quiz lacks the sampling controls needed for a population estimate. On Ars Technica, \citet{ars2026nblite} focused on Nano Banana 2 Lite's low latency and per-image cost. On Tom's Hardware, \citet{tomshardware2026qwen21} contrasted Alibaba's benchmark claims for Qwen-Image 2.1 with the need for independent testing. Following the Google Earth rollback, editorial commentary examined the risks of embedding image generation inside a trusted mapping tool \citep{techradar2026ge,pcmag2026nbge}. Across outlets, these reviews weigh vendor positioning against practical deployment risks.

\subsection{Developer-facing resources}
Developer documentation clarifies operational constraints around inputs, licensing, and provenance inspection. API reference pages from \citet{qwen2026api30} specify reference-image limits for hosted Qwen-Image 3.0 endpoints. Licensing terms from \citet{ideogram2026license} grant Apache 2.0 rights for Ideogram 4.0 inference code and restrict model weights under a Non-Commercial License, so practitioners must assess code and weight permissions separately. Provenance guides and Google's open-source Credentio library document programmatic checks for supported C2PA manifests and watermarks \citep{openai2026provenance,google2026credentio}.

\subsection{What the discussion adds to the risk picture}
Practitioner threads, editorial reviews, and developer guides point to recurring evaluation questions around label accuracy, appeal mechanisms, unit pricing, license scope, and vendor benchmark claims. Keeping user complaints, journalist tests, and API specifications in separate evidentiary roles turns informal commentary into structured questions for empirical testing.

\begin{figure}[!htbp]
    \centering
    \includegraphics[width=0.94\textwidth]{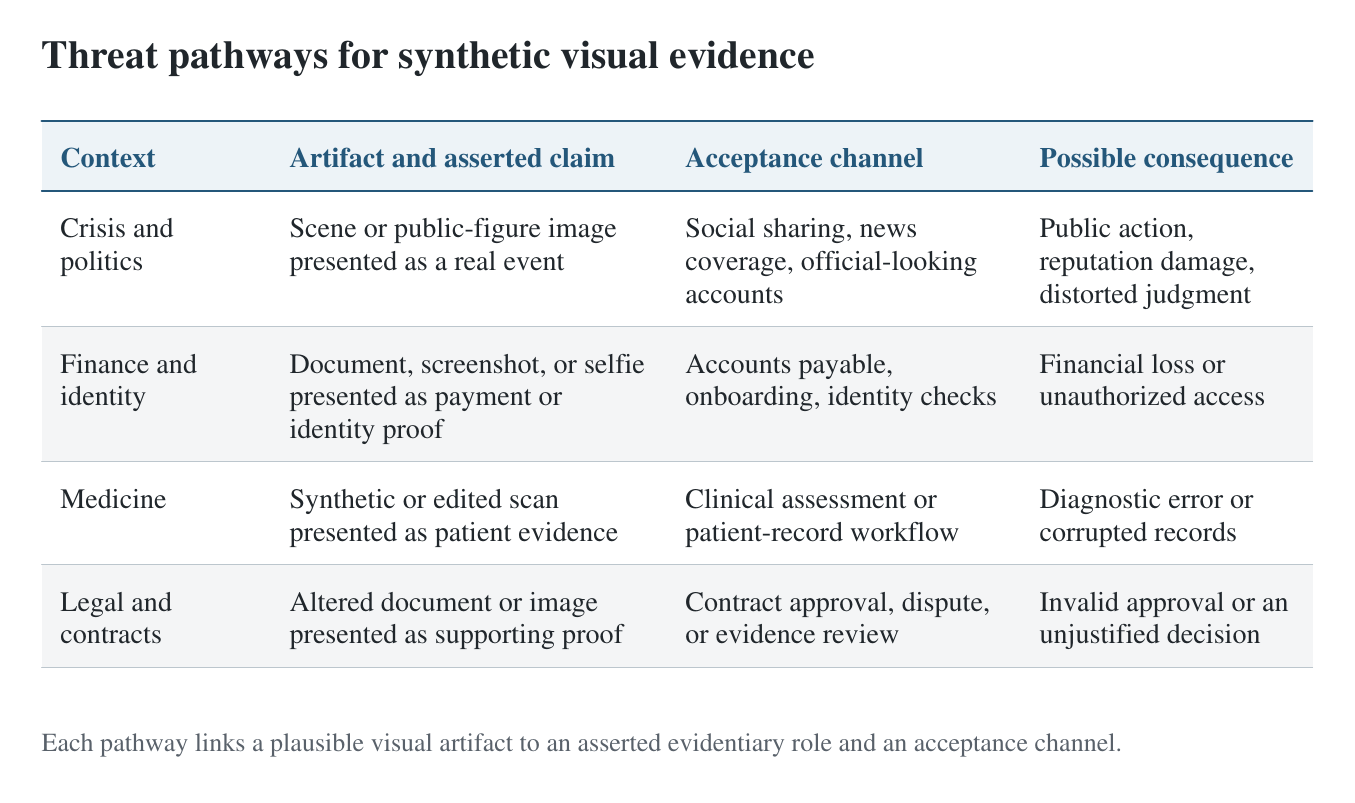}
    \caption{Threat pathways connecting synthetic artifacts, asserted claims, acceptance channels, and potential consequences across four institutional contexts.}
    \label{fig:risk_surface}
\end{figure}

\section{Risk Taxonomy}
\label{sec:risk_taxonomy}

{\small
\setlength{\tabcolsep}{4pt}
\begin{longtable}{>{\raggedright\arraybackslash}p{0.15\textwidth}>{\raggedright\arraybackslash}p{0.23\textwidth}>{\raggedright\arraybackslash}p{0.30\textwidth}>{\raggedright\arraybackslash}p{0.245\textwidth}}
\caption{Prospective harm pathways and institutional controls for synthetic visual evidence.}\label{tab:taxonomy}\\
\toprule
\textbf{Domain} & \textbf{Example artifact} & \textbf{Potential harm pathway} & \textbf{Suggested controls}\\\midrule
\endfirsthead
\multicolumn{4}{l}{\textbf{Table \thetable\ continued.}}\\
\toprule
\textbf{Domain} & \textbf{Example artifact} & \textbf{Potential harm pathway} & \textbf{Suggested controls}\\\midrule\endhead
\midrule\multicolumn{4}{r}{Continued on next page}\\\endfoot
\bottomrule\endlastfoot

Finance & Crisis photo, invoice, receipt, payment screenshot, identity document & Market panic, misdirected payments, account takeover, false compliance evidence & Direct callback, signed transaction logs, provenance-aware document intake, multi-person approval \\
Medicine & X-ray, scan, lab image, doctor note, insurance image & Incorrect treatment, fraudulent claims, litigation manipulation, clinical confusion & Trusted capture chain, PACS audit, clinician review, image-source logging \\
News and public safety & Fire, explosion, flood, conflict, accident, protest image & Public panic, emergency misinformation, harmful amplification before verification & Crisis friction, geolocation, independent source confirmation, provenance display \\
Law and contracts & Contract page, seal, signature, letterhead, court notice, notary artifact & False evidence, forged authority, coercive demands, reputation damage & Document registry checks, digital signatures, attorney verification, chain-of-custody rules \\
Identity and KYC & ID card, passport, selfie, liveness image, badge & Account opening fraud, synthetic identity, credential laundering & Liveness plus device attestation, biometrics with human review, C2PA-aware capture \\
Civic discourse & Public-figure photo, celebrity image, campaign poster, fake screenshot & Defamation, voter manipulation, liar's dividend, harassment & Public-figure protections, platform labels, newsroom correction channels \\
Enterprise security & E-mail image, fake ticket, badge, UI screenshot, internal memo & Social engineering, credential theft, insider impersonation & Security training, ticket validation, domain-bound workflow checks \\
\end{longtable}
\par\footnotesize\textit{Note.} Entries pair plausible harm mechanisms with institutional controls rather than empirical incident rates or comparative control efficacy. KYC denotes know-your-customer verification, and PACS denotes a clinical picture archiving and communication system.
}

\subsection{Why realism alone does not explain risk}
Pixel-level photorealism accounts for only one element of evidentiary risk. Documented episodes also rely on publication venue, recognizable identities, institutional formatting, or urgent demands. Even a visually imperfect image can trigger a costly error if a recipient defers to the apparent authority of its sender or host interface. Risk assessment must therefore evaluate the artifact alongside its presentation context and the decision it seeks to prompt.

\subsection{Capability interactions}
Different misuse pathways combine realism, legible typography, reference consistency, localized editing, search grounding, and low production cost in task-specific ways. Crisp text and page layout shape counterfeit invoice or notice plausibility \citep{openai2026images25,qwen2026api30}, multi-reference identity persistence supports impersonation \citep{google2025nanobananapro,qwen2026image21}, and low-cost tiers reduce barriers to repeated drafting \citep{google2026nanobanana2lite}. Operational evaluations should test these properties jointly within the target decision workflow.

\subsection{The verification lag}
\begin{figure}[!htbp]
    \centering
    \includegraphics[width=0.86\textwidth]{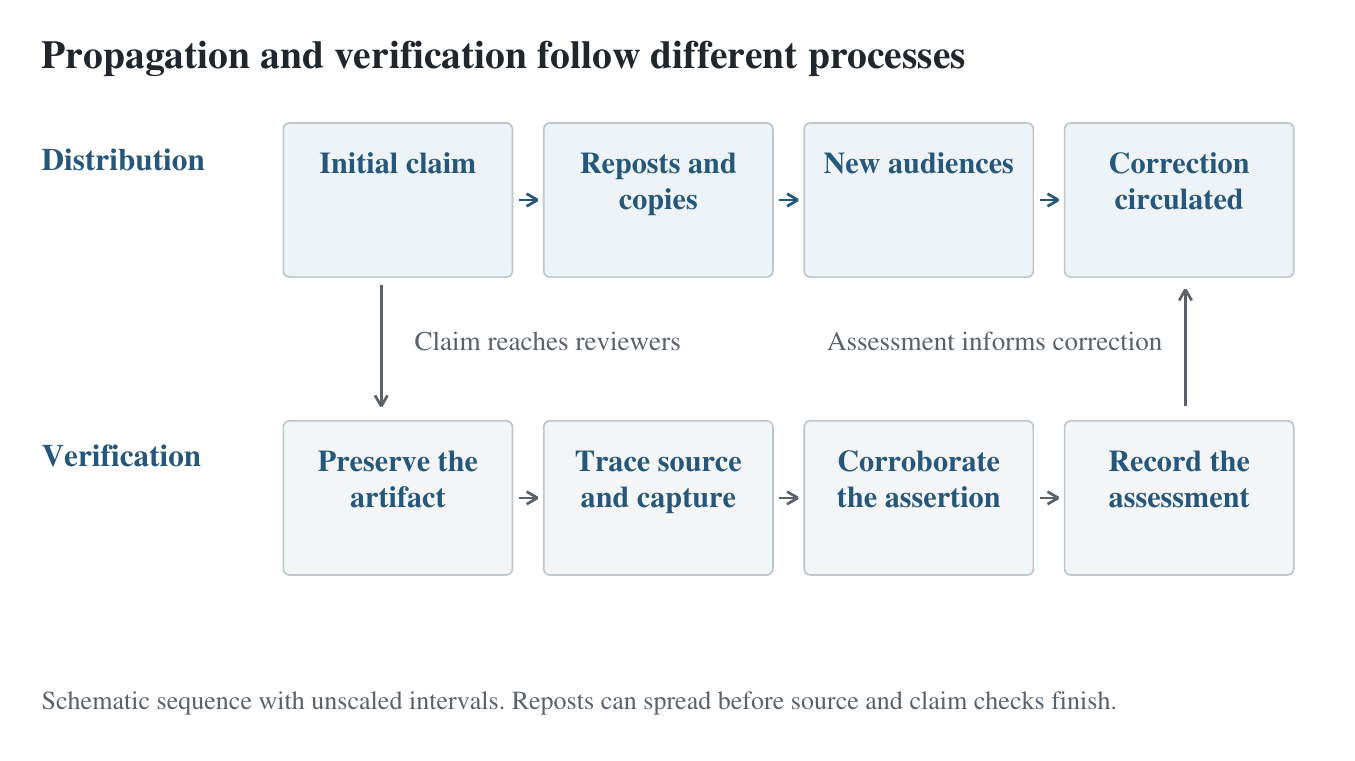}
    \caption{Sequence of publication, amplification, verification, and correction. Unscaled stages show where consequential action can precede corroboration rather than measuring durations or audience-response rates.}
    \label{fig:verification_lag}
\end{figure}

Online distribution routinely outpaces journalistic and institutional verification. During the May 2023 Pentagon episode, a fabricated explosion image spread across social feeds and coincided with brief market movement before official denials took hold \citep{ap2023pentagon,reuters2023pentagon}. Figure~\ref{fig:verification_lag} contrasts these parallel processes and marks the window in which audiences or institutions may act on an unverified visual claim.

\subsection{Two-sided inference problems}
Origin disclosure, production method, and factual truth present distinct questions. In the Chongqing bridge-collapse case, an explicit AI-generated label accompanied a harmful false emergency claim \citep{cq2026labelcase}. Conversely, an image lacking a recoverable watermark may be either an authentic camera capture or a synthetic output from an unwatermarked or re-encoded source. Provenance documentation from \citet{openai2026provenance} restricts verification results to supported origin signals and excludes factual accuracy or contextual truth. High-stakes review must therefore combine origin inspection with independent checks of the depicted event or transaction.

\begin{figure}[!htbp]
    \centering
    \includegraphics[width=0.93\textwidth]{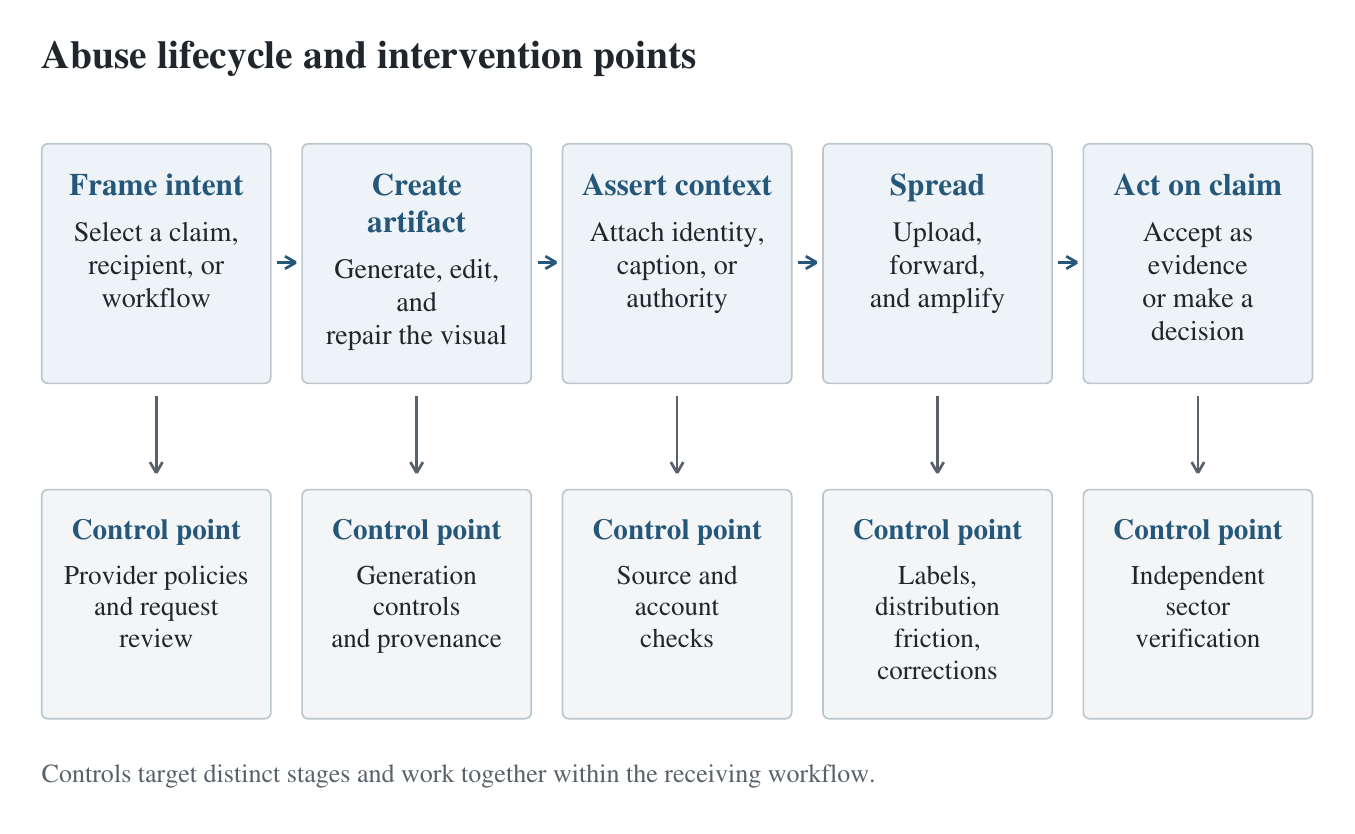}
    \caption{Lifecycle of synthetic visual misuse and defensive intervention points. Individual incidents can enter or bypass stages depending on the distribution channel.}
    \label{fig:abuse_lifecycle}
\end{figure}

\begin{figure}[!htbp]
    \centering
    \includegraphics[width=0.92\textwidth]{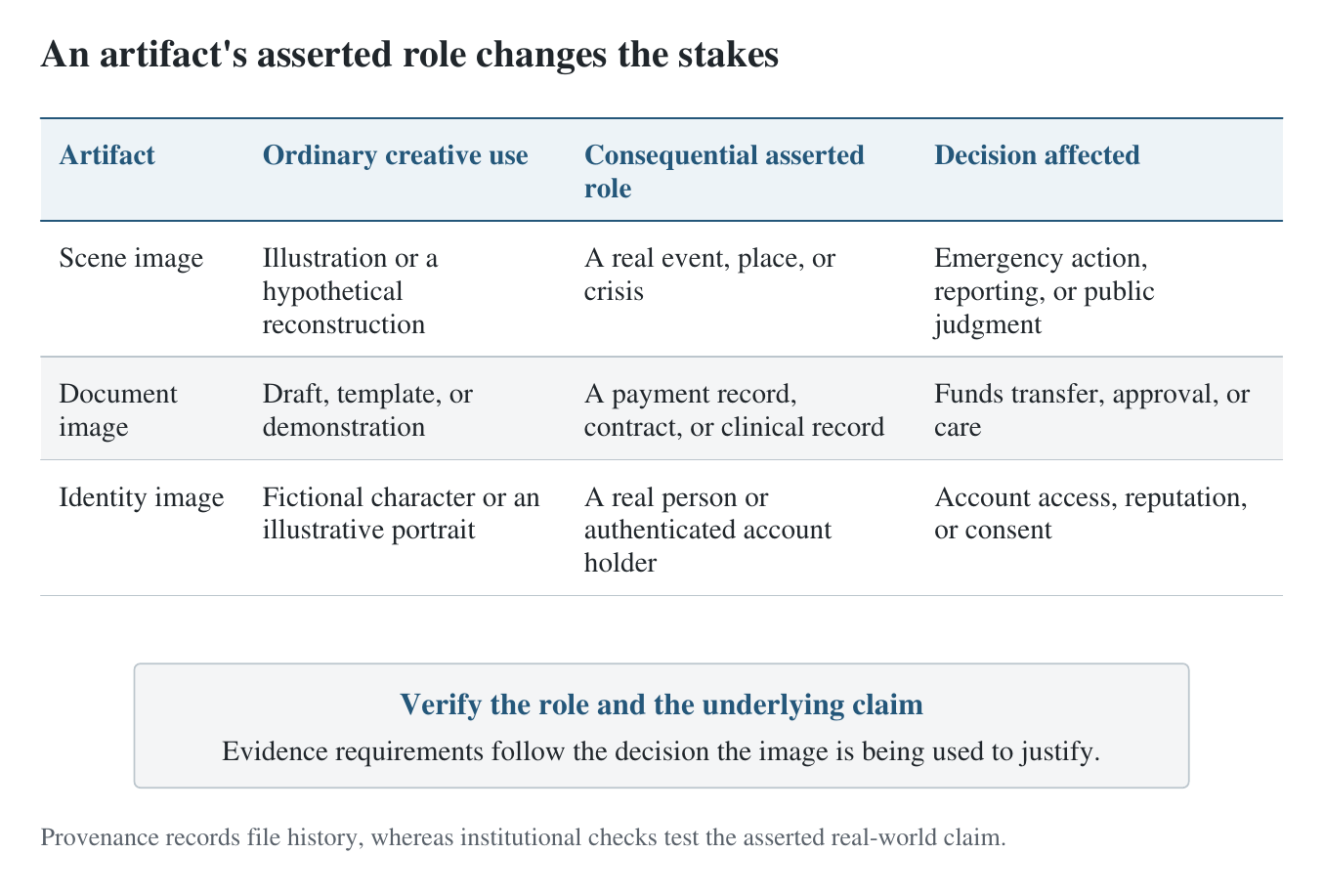}
    \caption{Evidentiary roles assigned to synthetic images in institutional workflows, linking asserted meaning to downstream decision stakes.}
    \label{fig:domain_escalation}
\end{figure}

\section{Current Governance and Control Measures}
\label{sec:governance}

\subsection{Provider-level controls}
Hosted providers combine prompt filtering, output refusals, account restrictions, service logging, and provenance metadata. OpenAI documents safety guardrails and provenance credentials for ChatGPT Images \citep{openai2026images25,openai2026provenance}, and xAI developer release notes introduce a request-level safety identifier for API abuse attribution \citep{xai2026releasenotes}. Investigation findings from Canada's Privacy Commissioner record additional safeguards represented by X Corp. and xAI while reserving judgment on their operational effectiveness \citep{opc2026grok}. China's labeling rules assign explicit and implicit disclosure duties to covered generation and dissemination services \citep{cac2025labeling}. Local open-weight inference removes hosted account controls and shifts compliance with license terms and statutory law to the deploying operator and jurisdiction.

\subsection{Provenance, watermarking, and content credentials}
C2PA content credentials bind cryptographically signed origin and edit assertions to media files \citep{c2pa2026assertions}. OpenAI pairs C2PA manifests with SynthID watermarking and a verification service for supported outputs \citep{openai2026provenance}, and Google embeds SynthID across Nano Banana Pro and Nano Banana 2 Lite \citep{google2025nanobananapro,google2026nanobanana2lite}. Recovered credentials or watermarks supply evidence about file production, whereas the depicted event and accompanying caption still require independent corroboration. Deployers must also test whether routine re-encoding, cropping, or messaging transfers preserve those signals.

Implementation guides include an IPTC FAQ on C2PA credentials and Google's open-source Credentio C++ library \citep{iptc2026c2pafaq,google2026credentio}. Provenance reliability depends on trusted signing certificates, recorded assertion types, and downstream metadata retention. Receiving workflows should log verification outcomes alongside their coverage limits and corroborate the asserted event whenever a decision turns on what an image purports to show.

\subsection{Policy status}
Statutory and voluntary transparency rules took shape across several jurisdictions through October 2026. In the European Union, the European Commission published its final marking and labeling Code of Practice on 10 June 2026 and issued implementation guidelines on 20 July \citep{europeancommission2026code,ec2026codefinal,ec2026guidelines}. Binding Article 50 transparency obligations took effect on 2 August 2026, and adherence to the Code of Practice remains voluntary. China's measures on explicit and implicit synthetic-content labels took effect on 1 September 2025 for covered generation and distribution platforms \citep{cac2025labeling}. Second-stage reporting on a 2026 Chinese enforcement campaign recorded more than 5.6 million items of illegal information removed and about 49,000 accounts sanctioned across all campaign categories \citep{people2026qinglang}. Those totals span the full campaign rather than isolating AI-generated visuals. In May 2026, the U.S. Federal Trade Commission began enforcing the TAKE IT DOWN Act \citep{ftc2026takedown}. In the United Kingdom, a parliamentary research briefing surveyed domestic AI-labeling proposals \citep{ukcommons2026labelling}. Because these instruments differ in statutory reach, institutional compliance must be assessed against each measure's specific duties.

\begin{figure}[!htbp]
    \centering
    \includegraphics[width=0.83\textwidth]{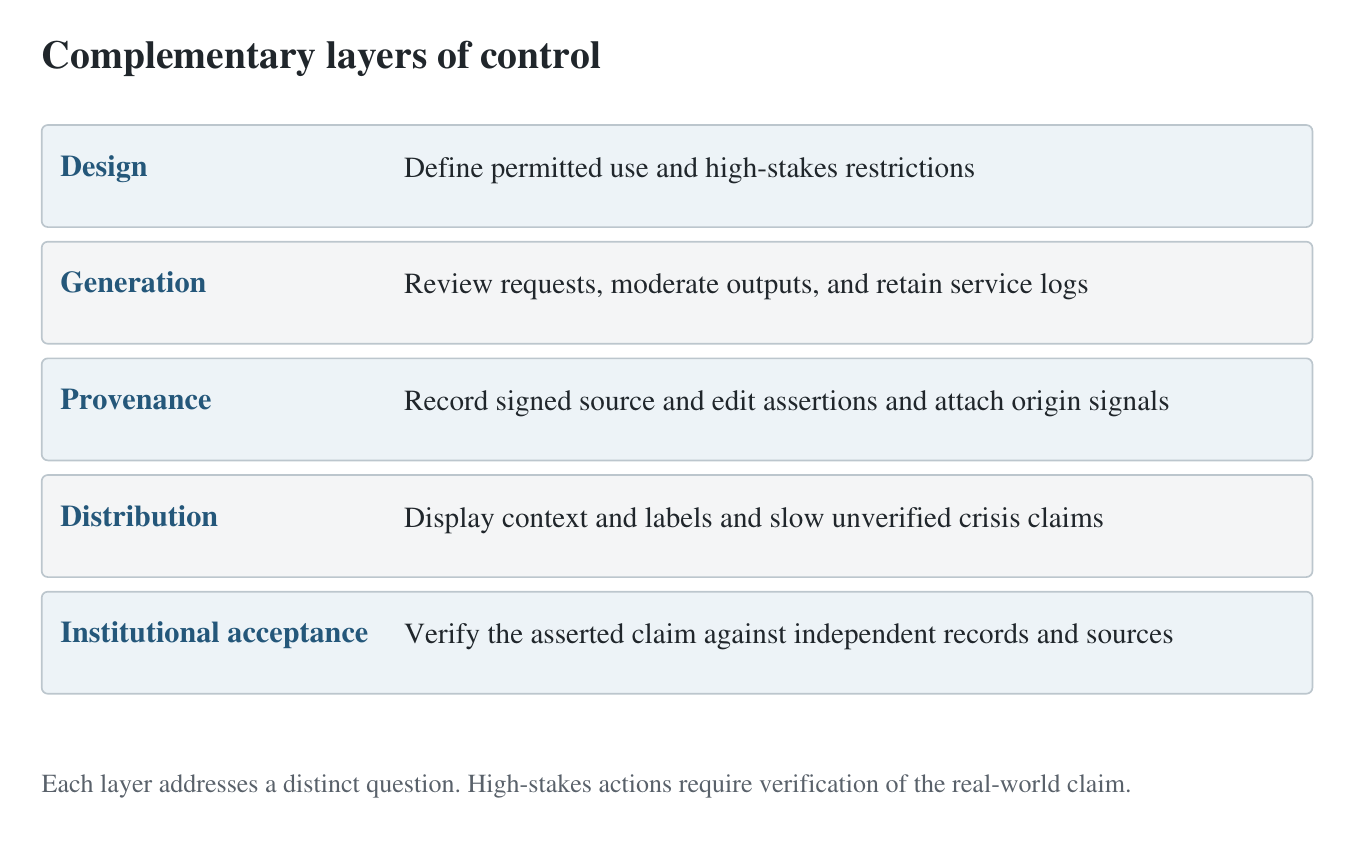}
    \caption{Complementary control layers for synthetic visual evidence. File-origin signals and independent corroboration of the asserted claim address distinct questions.}
    \label{fig:control_stack}
\end{figure}

\begin{figure}[!htbp]
    \centering
    \includegraphics[width=0.96\textwidth]{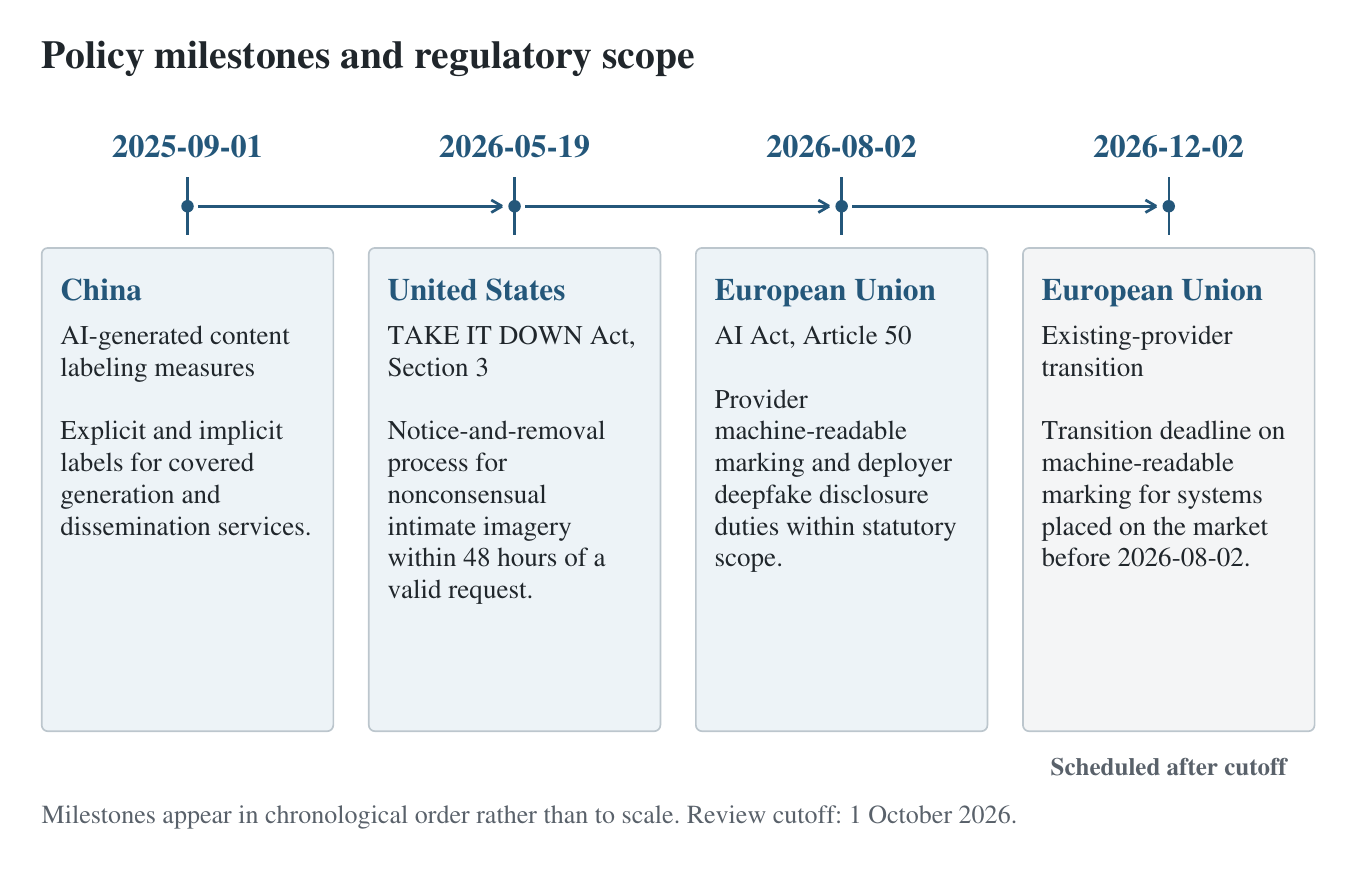}
    \caption{Policy implementation dates and regulatory scope in China, the United States, and the European Union \citep{cac2025labeling,ftc2026takedown,ec2026guidelines}. A 2 December 2026 EU transition deadline governs machine-readable marking for systems placed on the market before 2 August 2026 and falls after the 1 October 2026 review cutoff.}
    \label{fig:policy_maturity}
\end{figure}

\subsection{Platform labeling practice}
Platform disclosure rules depend on content category and available detection signals. Meta's advertising guidance separates photorealistic AI-generated people and substantive edits from minor creative-tool adjustments, applying disclosures from advertiser self-reports or industry provenance metadata \citep{meta2026ads}. On 27 May 2026, YouTube introduced more prominent disclosures for photorealistic or meaningfully altered content and began automatically labeling detected synthetic videos left undisclosed by uploaders \citep{youtube2026labels}. Platform policies define trigger rules and label placement rather than universal detection, so independent evaluations should test label coverage, viewer comprehension, and appeal mechanisms.

\subsection{Detection within a verification workflow}
Detector outputs require interpretation within the conditions under which each tool was validated. Using MarkNull, \citet{cao2026marknull} show that on-manifold latent manipulation can strip watermarks from AI-generated images under adversarial post-processing. Radiologist discrimination of synthetic radiographs likewise varies with task framing and prior warning \citep{tordjman2026xrays}. Receiving institutions should record the detection tool, input handling history, and confidence bounds before checking primary source records. Automated classification alone leaves the truth of a depicted event unresolved.

\begin{figure}[!htbp]
    \centering
    \includegraphics[width=0.93\textwidth]{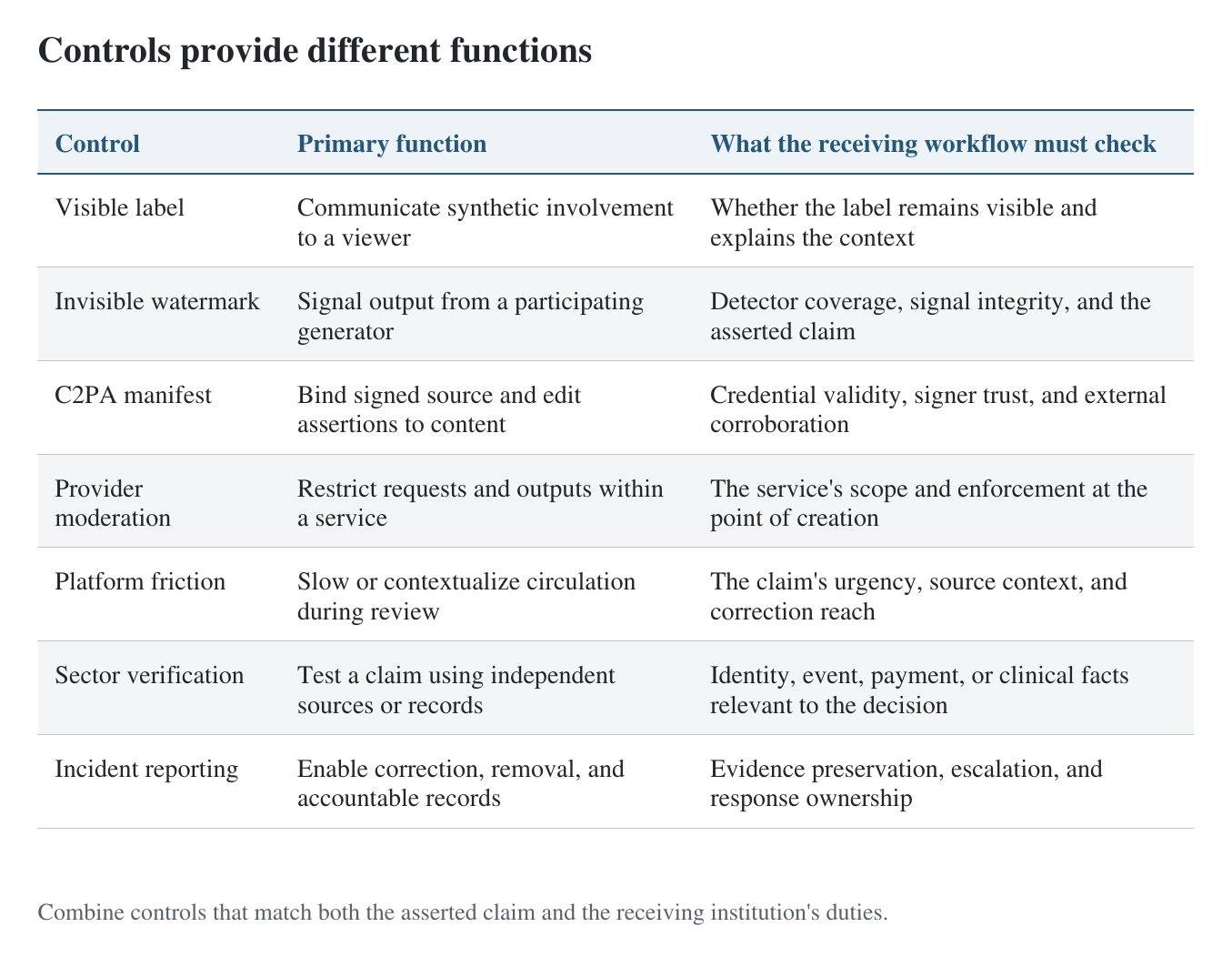}
    \caption{Functional comparison of control mechanisms. Each row pairs a control's primary check with the external verification required for high-stakes decisions.}
    \label{fig:mitigation_scorecard}
\end{figure}

\section{Recommendations}
\label{sec:recommendations}

\subsection{For model providers}
Model providers should scrutinize requests that generate artifacts suited to identity proofing, institutional authorization, or evidentiary proof. Priority categories cover identity cards, corporate notices, clinical scans, crisis scenes, and real-person likenesses. Baseline controls include default provenance credentials, context-appropriate disclosures, watermark-robustness testing, and targeted blocks on document forgery and non-consensual intimate imagery. High-risk API integrations also need accountable customer vetting and an accessible abuse-reporting channel for affected individuals and organizations.

Safety review should focus on requests capable of supporting consequential deception so that clearly framed creative and educational uses remain practical. Economy and batch tiers warrant the same abuse testing as flagship models because low unit pricing facilitates iterative refinement and scale. Public transparency reports should disclose confirmed abuse episodes, tested refusal categories, and provenance survival rates with explicit denominators so that external reviewers can separate measured outcomes from policy pledges.

\subsection{For platforms and newsrooms}
Distribution platforms should surface readable provenance metadata when present and attach corrections directly to debunked posts. During breaking emergencies, source-attribution prompts and temporary friction on unverified visual claims can buy time for fact-checking, and platforms should evaluate their operational impact. Mapping and archival products that incorporate generative features must visually separate optical records from synthetic reconstructions \citep{google2026earth,pcmag2026nbge}. Newsrooms should pair chain-of-origin checks with reverse image search, geolocation, direct outreach to local witnesses, and prominent public corrections.

\begin{figure}[!htbp]
    \centering
    \includegraphics[width=0.96\textwidth]{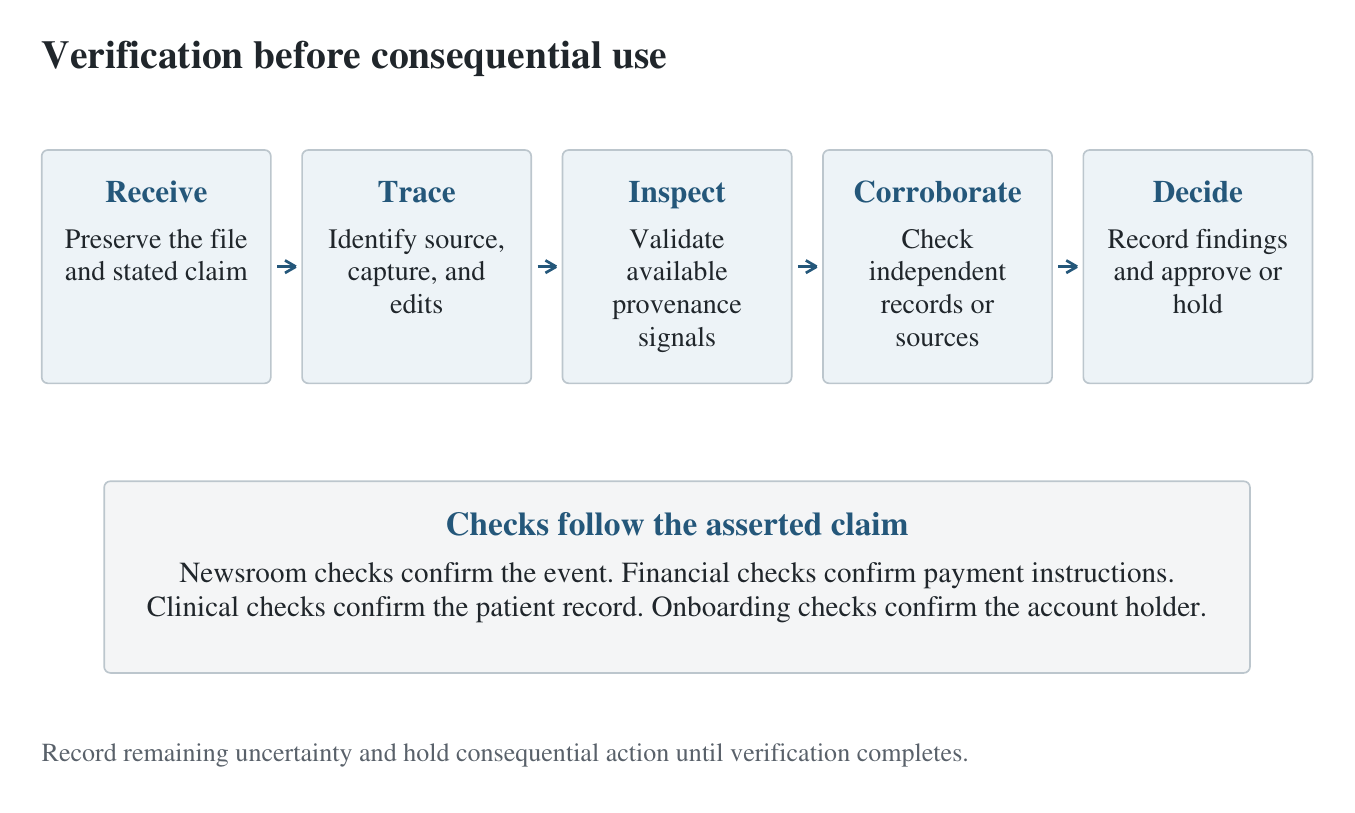}
    \caption{Verification workflow for high-stakes visual claims. Provenance inspection checks file origin and edit history, and independent source checks test the depicted event, document, or transaction.}
    \label{fig:verification_workflow}
\end{figure}

\subsection{For financial institutions and enterprises}
Banks, accounting departments, and enterprises should require authenticated primary records before executing high-stakes actions. Staff should confirm changes to payment instructions via independently verified callback numbers, reconcile transfer claims against ledger systems, and route exceptions through dual authorization. Screenshots and formatted PDFs may initiate a request, yet final approval must rest on verified transaction records and authenticated requester authority.

\subsection{For healthcare and legal systems}
Healthcare organizations should maintain cryptographic or logged links connecting each diagnostic image to its capture modality, patient record, and subsequent processing. Externally uploaded or referred scans warrant origin checks and clinical correlation before guiding treatment or reimbursement. Courts and legal practitioners should preserve original digital files and chain-of-custody logs, pairing provenance inspection and forensic analysis with witness testimony and documentary records. Content credentials document aspects of file handling, and evidentiary weight turns on the factual claim under adjudication.

\subsection{For regulators}
Regulators should define clear statutory boundaries for synthetic-media disclosures and publish operational compliance guidance for covered services. Audits and incident reporting can verify whether mandated labels persist in practice and whether takedown or appeal procedures function as stated. Regulatory evaluation should track both technical compliance and downstream harm reduction across identity verification, emergency alerts, financial authorization, clinical records, and judicial evidence.

\subsection{For ordinary users}
Individual viewers can check who first posted an image, whether independent outlets corroborate the event, and whether the named institution records the notice or transaction on its official channels. Available provenance tools can clarify whether a supported AI service produced or edited the file. Demands involving wire transfers, urgent action, or secrecy warrant confirmation over an independently verified contact channel before sharing or compliance.

\begin{figure}[!htbp]
    \centering
    \includegraphics[width=0.88\textwidth]{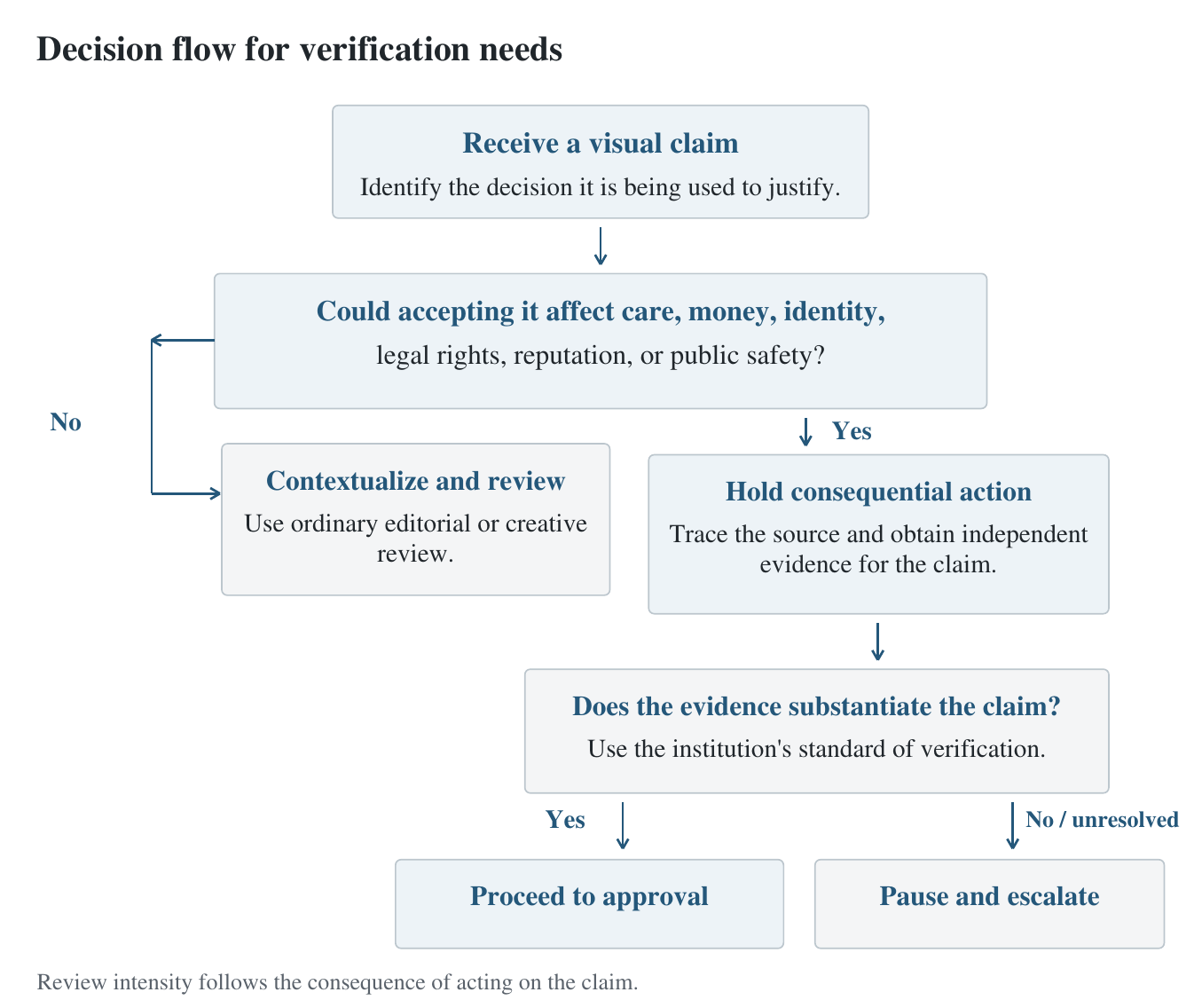}
    \caption{Decision flow matching verification rigor to the stakes of acting on a visual claim and the strength of independent corroboration.}
    \label{fig:priority_map}
\end{figure}

\section{Discussion}
\label{sec:discussion}

\subsection{Verification of consequential visual claims}
Our framework centers verification on the institutional decision an image is used to justify. Banks reconcile transfer claims against settlement ledgers, hospitals match diagnostic images against modality and PACS logs, and newsrooms corroborate crisis visuals with independent witnesses on the ground. Provenance credentials and detectors can flag synthetic origin or editing, yet the receiving institution must set corroboration thresholds proportionate to the cost of a wrong decision.

\subsection{The hard case of mixed media}
Real-world artifacts frequently blend optical capture, conventional editing, and generative synthesis. An authentic camera photograph may undergo localized AI inpainting. Likewise, a genuine paper contract may be composited into a staged scene, or an unaltered interface screenshot may circulate under a false caption. Useful disclosures therefore specify which capture or editing steps occurred when provenance metadata survives. Even complete edit histories leave recipients to verify whether the surrounding claim represents the artifact accurately.

\subsection{Questions raised by public discussion}
Practitioner threads point to operational friction around false labels and appeal routes, and independent reviews question vendor benchmark claims that lack open test protocols. Treating public complaints and editorial tests as design hypotheses helps platforms and institutions audit label clarity, appeal turnaround, and staff verification procedures in their own workflows.

\subsection{International asymmetry}
Statutory obligations differ across jurisdictions in how they allocate duties among model providers, API deployers, platforms, and publishers. Cross-border forwarding routinely strips metadata and obscures legal accountability. Interoperable C2PA credentials and standardized intake logs can reduce friction across jurisdictions. Open-weight local inference shifts technical enforcement away from hosted APIs, yet operators remain bound by model licenses and national laws governing fraud, defamation, privacy, and synthetic disclosures.

\subsection{Beneficial uses and proportionate controls}
Image generators support commercial design, educational illustration, accessibility, and explicitly labeled historical or spatial reconstruction. Proportionate governance evaluates how an artifact is framed and used, concentrating scrutiny on real-person or institutional impersonation and on workflows where consequential decisions rely on visual proof. Matching verification rigor to decision stakes preserves benign creative work and hardens identity, financial, clinical, and legal intake channels.

\section{Limitations}
\label{sec:limitations}

This narrative review synthesizes public documentation and cannot estimate undetected private abuse or population-wide prevalence. Vendor release notes describe advertised capabilities rather than head-to-head performance or safety under shared benchmarks, and public hoax reports rarely confirm which model generated an artifact. Referenced English and Chinese community threads reflect self-selected contributors and uneven platform retrieval, providing qualitative context rather than public-opinion estimates. Similarly, our risk taxonomy links capabilities to plausible workflow failures and controls without estimating causal effect sizes. Cited experimental findings on synthetic radiographs and commercial fraud shares from Entrust remain bounded by their respective reader cohorts, customer populations, and collection windows. Subsequent model releases, watermark tools, and regulatory enforcement after 1 October 2026 may alter specific capability and compliance details.

\section{Conclusion}
\label{sec:conclusion}

Contemporary image generators combine photorealistic synthesis with legible multilingual text, structured layouts, multi-reference identity persistence, search grounding, and low-cost iterative editing. Documented public episodes show synthetic visuals circulating as crisis photography, public-figure encounters, and administrative notices. Controlled radiograph reader experiments and commercial document-fraud statistics add bounded empirical evidence within clinical reading and onboarding settings. Together, these public sources clarify how generated or edited visuals can acquire apparent evidentiary authority inside institutional workflows.

Effective defense shifts emphasis from visual inspection alone to verification of the claim an image is used to support. Model providers can attach durable provenance signals and restrict high-risk impersonation or document forgery. Platforms can keep disclosures and corrections attached to viral claims, and high-stakes institutions can require authenticated primary records and out-of-band corroboration before acting. Provenance metadata documents file production and editing history, whereas verifying a depicted event or transaction requires independent source checks. Future empirical work should test these layered workflows on realistic institutional tasks and measure error rates, latency, and operational resilience.

\section*{Data and Code Availability}
The accompanying source package includes the \LaTeX{} manuscript, BibTeX database, Python figure-generation script, vector figure PDFs, style files, and structured research notes supporting the narrative synthesis. Running the figure script reproduces all conceptual diagrams and the Entrust fraud-category chart from published figures rather than an original experimental dataset.

\section*{Acknowledgments}
The authors thank Xinwei Du for support and Yanna Feng for guidance and academic advice during the preparation of this manuscript.

\FloatBarrier
\begingroup
\interlinepenalty=10000
\bibliography{references}
\endgroup

\end{document}